\documentclass{article}

\usepackage[position, final]{neurips_2026}

\usepackage[utf8]{inputenc} 
\usepackage[T1]{fontenc}    
\usepackage{hyperref}       
\usepackage{url}            
\usepackage{booktabs}       
\usepackage{amsfonts}       
\usepackage{nicefrac}       
\usepackage{microtype}      
\usepackage{xcolor}         

\usepackage{comment}
\usepackage{graphicx}
\usepackage{amsmath}
\usepackage{multirow}
\usepackage{enumitem}
\usepackage{caption}
\usepackage{wrapfig}
\usepackage{enumitem}
\usepackage{placeins}

\title{AI Evaluation Should Require Standardized Item-Level Data Releases}

\author{
  Han Jiang$^{1}$ \quad Susu Zhang$^{2}$ \quad Dongyao Zhu$^{5}$ \quad Yuzhuo Bai$^{6}$ \quad Sang T. Truong $^{4}$ \\
  \quad \bfseries Xiaoyuan Yi$^{3}$ \quad Sanmi Koyejo$^{4}$ \quad Xing Xie$^{3}$ \quad Ziang Xiao$^{1}$\\[0.5em]
  $^1$Johns Hopkins University \quad $^2$University of Illinois Urbana-Champaign \\
  $^3$Microsoft Research Asia \quad $^4$Stanford University \quad $^5$North Carolina State University \\
  $^6$Tsinghua University\\
  \texttt{hjiang66@jh.edu, ziang.xiao@jhu.edu}
}

\begin{document}

\maketitle

\begin{abstract}
This position paper argues that standardized item-level benchmark data should become the default infrastructure for AI evaluation.
Current evaluations suffer from underspecified item selection, construct misalignment, and poor generalization;
a disproportionate focus on aggregate model scores compounds these failures by keeping them undetected and unfixable.
Without item-level evidence, validity claims cannot be assessed, resulting in inflated capability claims, misdirected research, and unwarranted trust in deployed systems.
Our position is that valid evaluation requires empirical evidence from item-level benchmark data, including item content, test conditions, model responses, and item-level score statistics, and that the standardized release of such data should be treated as core AI evaluation infrastructure.
Such releases also enable transparency, replicability, and auditability of evaluation results.
To show that the norm is both feasible and consequential, we construct \textsc{OpenEval}, an item-level repository of 10.7M+ responses across 160K+ items from widely-used benchmarks, under a unified schema that the AI evaluation community can build upon.
We demonstrate how item-level data can identify low-quality items, document construct misalignment, and recover validity evidence about benchmarks' internal structure.
We address objections concerning contamination and coordination burden, and show that both are tractable relative to the costs of decisions made on inadequately validated evaluative claims.
\end{abstract}


\section{Introduction}\label{sec:1}
Generative AI is moving rapidly into high-stakes deployments, while AI evaluation, dominated by benchmarking practice ~\citep{Eriksson_Purificato_Noroozian_Vinagre_Chaslot_Gomez_Fernandez-Llorca_2025}, has become the primary instrument for understanding model capabilities, informing AI policy, and guiding responsible deployment.  
However, the empirical foundation of such instruments is thinner than their influence suggests.
Benchmarks are released, leaderboards are populated, and deployment decisions are made, while the underlying response data that would let the community audit whether benchmarks measure what they claim to measure and evaluate benchmark reliability are rarely shared and analyzed. 

This gap entails consequences. Critical design choices, including capability definitions, content curation, and metric selection, often lack transparency or formal justification~\citep{liu-etal-2024-ecbd}.
This opacity undermines the validity evidence~\citep{blodgett-etal-2021-stereotyping,liu-etal-2024-ecbd} needed to support the interpretations of results, making it unclear whether benchmarks genuinely measure their intended constructs, despite explicit metadata or task descriptions ~\citep{akyurek-etal-2022-challenges}. Compounding these design limitations, the speed of AI development manifests as benchmark saturation~\citep{ott2022benchmark}, rapidly outdated content~\citep{jiang2025time}, and widespread data contamination~\citep{xu-etal-2025-dcr}, rendering aggregate scores uninformative or misleading for deployment decisions~\citep{golchin2024time}. 
New benchmarks proliferate to address these concerns, but often recycle existing practices without adding evaluative insight~\citep{blodgett-etal-2020-language}, widening the socio-technical gap between technical solutions and real-world requirements~\citep{10.5555/3042573.3042809,liao2025rethinking}.

\begin{wrapfigure}{r}{0.5\textwidth}
    \vspace{-15pt}
    \centering
    \includegraphics[width=\linewidth]{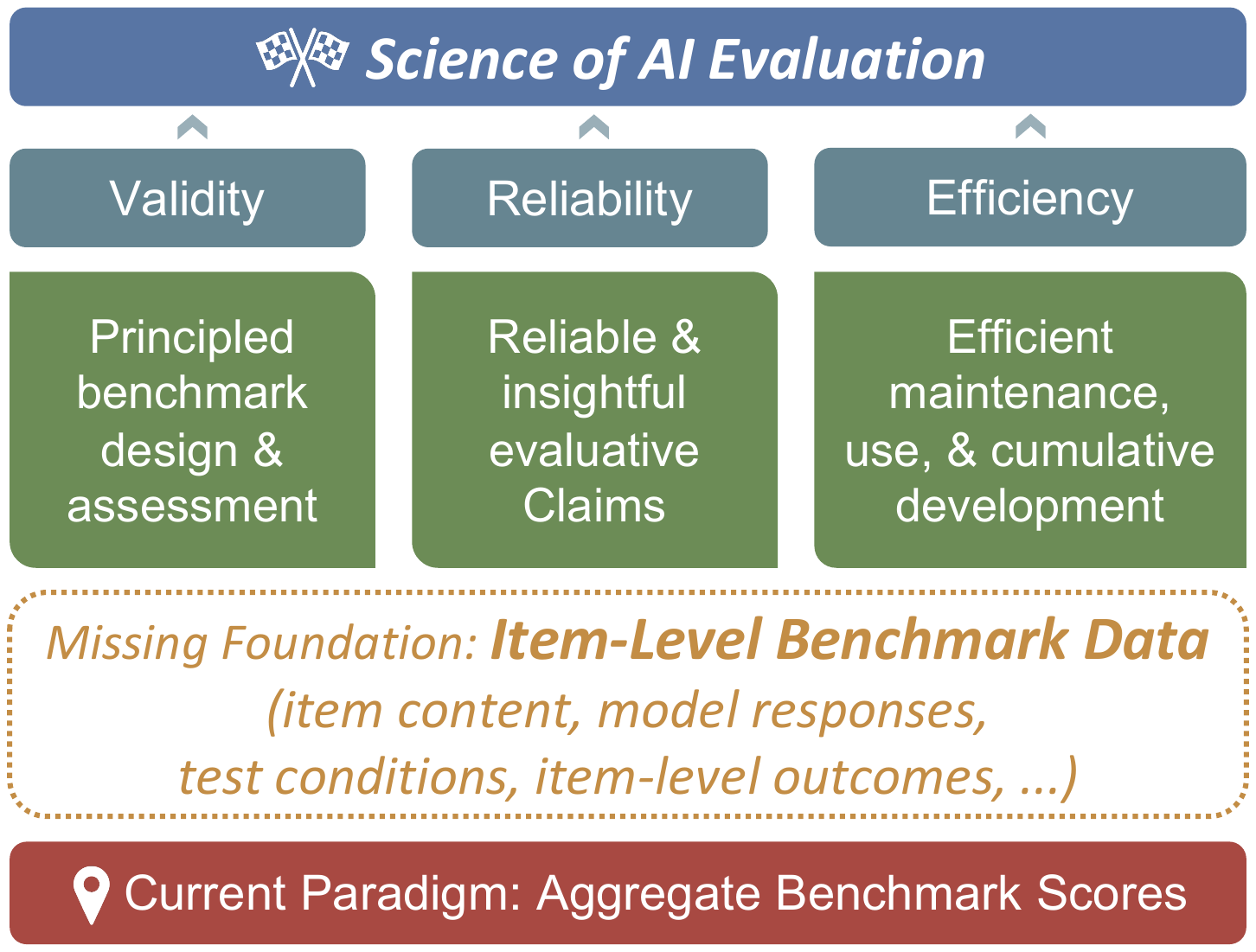}
    \caption{Illustrative overview of our position. Item-level benchmark data serves as the missing foundation between the current aggregate paradigm and the science of AI evaluation.}
    \label{fig:position}
\end{wrapfigure}

Crucially, many validity issues are not diagnosable from benchmark-level aggregate scores alone. Foundational questions---including whether items effectively differentiate model capabilities, how construct-irrelevant nuisance factors drive performance, or whether gains reflect genuine improvement rather than artifacts---are inherently item-level inquiries. Without item-level benchmark data, our field lacks the empirical evidence required to evaluate and curate effective benchmarks. 
We argue that \textbf{standardized item-level benchmark data should become the default infrastructure for AI evaluation, released alongside aggregate scores whenever legally and ethically feasible}: designing valid evaluations requires empirical evidence from item-level model responses, and the standardized release of such data should be treated as core evaluation infrastructure on par with the benchmarks themselves.

Shifting the focus from aggregate model scores to item-level model outputs has deep roots in measurement science, where item-level assessment data have long been significant to test development and validation across education, psychology, and other social science disciplines.
Bringing this granularity to AI evaluation would transform fragmented leaderboard results into cumulative empirical evidence: item content, score statistics, and per-item responses jointly enable rigorous construct validation, reliability analysis through item consistency and discrimination patterns, and efficiency improvements through identification of redundant or uninformative items. 
Item-level data is the foundation for AI evaluation science.

To show the norm is feasible and consequential, we construct \textsc{OpenEval}\footnote{Home page: \url{https://open-eval.com/}}, an item-level archive of 10.7M+ responses across 160K+ items from widely-used benchmarks, organized under a unified schema the community can contribute to. 
We use OpenEval to demonstrate how item-level data identifies low-quality items, documents construct misalignment, and recovers validity evidence about benchmarks' internal structure. 
At the end, we discuss broader opportunities and address objections around storage, contamination, and author burden.

\section{Validity Challenges in AI Benchmarking}\label{sec:2}
Despite their pivotal role, AI benchmarks face validity challenges from two directions, namely internal methodological limitations and external pressures from the rapid evolution of AI systems, both of which are exacerbated by the absence of item-level data.

\subsection{Methodological Issues in Benchmark Design}~\label{sec:2.1}
The term \textit{Validity}, specifically \textit{Construct Validity}, was formalized by Cronbach and Meehl~\citep{cronbach1955construct} in modern psychometrics and is directly applicable to AI evaluation~\citep{xiao2023evaluating}.
A \textit{Construct} refers to a theoretical, unobservable attribute or trait (e.g., a perceived AI capability) that a test is intended to measure. Validity thus reflects how well an assessment measures the intended underlying outcome~\citep{10.1007/978-3-030-29516-5_28} and is crucial for ensuring benchmark quality.

As noted in \citep{westen2003quantifying}, construct validity is a central concept in psychology, particularly for informing the design of psychological measures. 
In AI evaluation, however, validity has received limited attention, despite AI capabilities rapidly expanding and benchmark creation involving numerous design decisions, such as target capability identification, construct-to-task operationalization, item adaptation for AI systems, and metric selection. 
These decisions are often oversimplified due to \textit{a dominant focus on benchmark-level results}, leaving only standard or default settings (e.g., \citep{liang2023holistic, 10.5555/3454287.3454581}) and insufficient evidence for validity justification~\citep{liu-etal-2024-ecbd}.
These ambiguities arise when benchmark documentation leaves construct definitions implicit and provides limited item-level evidence supporting the intended score interpretations, making it difficult to articulate and communicate validity concerns consistently, highlighting the need for shared frameworks and guidelines for AI evaluation.
Although there are several studies exploring validity-relevant properties~(e.g., \citep{yao2025the,wu2025mapping}), research explicitly assessing AI benchmark validity remains limited. 
Existing analyses suggest that most benchmarks' validity remains underdeveloped and call for evidence beyond the benchmark level (e.g., \citep{blodgett-etal-2021-stereotyping,bean2025measuring,salaudeen2025measurement}).

Benchmarks designed with validity issues act as traps: their claimed goals may sound aligned with users' intended purposes, yet the underlying constructs being measured could diverge substantially, yielding unreliable conclusions, as demonstrated by~\citep{bean2025measuring}, that propagate through benchmark selection and all subsequent evaluation practices. 
Such misalignment can be further exacerbated by confounders unrelated to the intended system ability~\citep{xu-etal-2025-diagnosing}, such as erroneous items, spurious correlations, or unintended shortcuts~(e.g., \citep{du-etal-2021-towards,nahum-etal-2025-llms}). 

Notably, \textit{both construct misalignment and these overlooked confounders are only visible at the item level}, where the content of each item can be scrutinized and its empirical behavior systematically examined. 
The absence of item-level data hinders reasoning about what actually drives benchmark performance --- a question left unresolved from the design stage --- which not only wastes the effort invested in benchmark selection, but also leaves the validity of any downstream claim unjustified. 
Yet existing benchmark data remains largely untapped at this level, leaving benchmark design without the empirical feedback needed to become more principled.

\subsection{Pressures from AI Advances on Benchmarking Practices}~\label{sec:2.2}
Beyond the design limitations, the rapid evolution and increasing opacity of AI systems impose external pressures that further erode benchmark validity over time.

\begin{wrapfigure}{l}{0.5\textwidth}
    \vspace{-15pt}
    \centering
    \includegraphics[width=\linewidth]{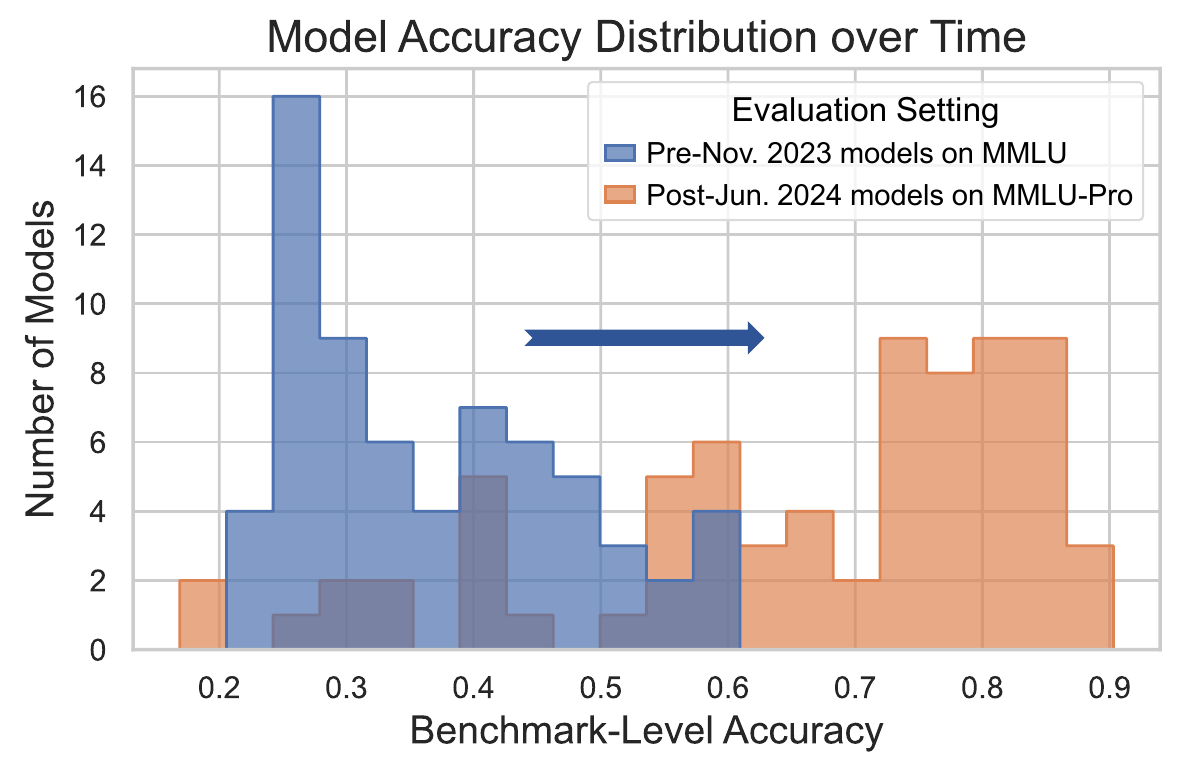}
    \caption{Benchmark-level accuracy distributions for 66 pre–Nov. 2023 models on \textsc{MMLU} and 72 post–Jun. 2024 models on \textsc{MMLU-Pro}. }
    \label{fig:acc}
    \vspace{-15pt}
\end{wrapfigure}

As AI systems and real-world knowledge co-evolve, benchmarks are subject to various forms of validity degradation~\cite{dehghani2021benchmarklottery,lifecycle}.

\textit{Benchmark saturation}. Some benchmarks have gradually become too easy to distinguish between the latest models~\citep{ott2022benchmark,deveci2025the}. 
For instance, \textsc{RealToxicityPrompts}~\citep{gehman-etal-2020-realtoxicityprompts} is one of the most widely used AI toxicity benchmarks; although by 2023 it could no longer differentiate between multiple versions of GPT~\citep{jiang2025raising}, it remained widely used beyond 2024. 

\textit{Outdated knowledge}. Benchmarks requiring factual knowledge are particularly time-sensitive, as outdated references render the corresponding test items unreliable~\citep{jiang2025time}.

\textit{Data contamination.} The opacity and information asymmetry in AI system development have made the risk increasingly pervasive.
As AI training and evaluation scale, many omnibus benchmarks further aggregate multiple previous benchmarks, making data provenance harder to trace.
Both intentional and inadvertent test-train overlap can secretly lead to unfair evaluations unless explicitly reported by developers~\citep{zhang2025position}.

\textit{All these issues are nearly impossible to detect at benchmark level}, and Fig.~\ref{fig:acc} exemplifies this diagnostic gap: the benchmark-level accuracy distribution has shifted markedly rightward as newer models are evaluated, even though \textsc{MMLU-Pro} was designed to be more challenging. Without item-level inspection, it remains unknown whether the observed improvements reflect genuine capability gains, benchmark saturation, or data contamination, \textit{leaving no valid claims about model capabilities to be drawn}.

This progressive degradation at scale creates a fundamental tension between evaluation efficiency and validity.
Manual benchmark updates~(e.g., \citep{lin2024wildbench,white2025livebench}) are believed to generate higher-quality items but are typically time-consuming and costly.
LLM-powered benchmarking~(e.g., \citep{kiela-etal-2021-dynabench,fadillamir2025benchmarking}) improves efficiency but at the expense of validity, as the quality of synthesized test items has been questioned in~\citep{bowman-dahl-2021-will}. 
More recent benchmark generation techniques, such as adversarial filtering~(e.g., \citep{10.5555/3524938.3525039,nie-etal-2020-adversarial}), item generation~(e.g., \citep{lin2024idgen,kim-etal-2025-stair}), and adaptive difficulty adjustment~(e.g., \citep{truong2025reliable}), offer promising directions but require exploring inter-item dynamics and engagement with measurement theories and psychometric methods.
Shared item-level data can support further development of dynamic benchmarking and automated item generation by enabling researchers to identify contaminated or outdated items, characterize the constructs they measure, and guide item retention, replacement, and generation as AI systems evolve.

Additionally, a common bottleneck underlying all of these efforts is the lack of shared data infrastructure. 
Current evaluation results are aggregated locally and not consistently released across leaderboards, making them neither comparable nor easily built upon, and forcing each study to independently assemble and preprocess its data. 
This duplicates effort across research groups and prevents cumulative progress. 
Moreover, ensuring the long-term sustainability of a unified item-level data release is itself a real challenge, as the effort required to curate, standardize, and update data across diverse benchmarks is prohibitive for any single research group.
That is to say, without \textit{a community-driven, consistent release of item-level benchmark data}, individual efforts to counter these ecosystem pressures remain isolated and unscalable.
\section{Item-Level Data: The Missing Foundation for AI Evaluation Science}\label{sec:3}
Given the twofold validity challenges in AI benchmarking and the limits of existing benchmarking practices, \textbf{item-level benchmark data is a crucial missing piece in the science of AI evaluation}.
Every benchmark run already generates such data: detailed test conditions, the content of each item, the model's response, and per-response scores and statistics. However, this evidence is routinely discarded once aggregate scores are computed. What remains is a rich yet largely inaccessible foundation for investigating the validity, reliability, and efficiency of AI benchmarking. We argue that releasing item-level data under a unified schema (e.g., Fig.~\ref{fig:schema}) should become a community norm when publishing evaluation results. Reconstituting this evidence into shared infrastructure is a precondition for evaluation to function as a cumulative and open science.

\textbf{Item-level data enables principled benchmark design and assessment.}
Item-level data provides an empirical foundation for developing and adapting psychometric methods for AI evaluation.
Existing models face challenges in this setting, including limited model samples, violations of statistical assumptions, and distribution shifts as AI systems evolve.
Shared item-level evidence enables researchers to examine these limitations and develop measurement models suited to AI evaluation.
Through more principled practices such as quantifying item characteristics, identifying decisive benchmark factors, and modeling relationships between items and intended constructs, future benchmarks can better clarify the mapping between AI competencies and evaluation objectives.

\textbf{Item-level data enables reliable and insightful evaluative claims.} 
Item-level data facilitates a bidirectional attribution that addresses the opacity in AI evaluation caused by the aggregate-focused paradigm.
First, observed model performance can be traced to measurable factors, disentangling intended constructs from confounders such as construct-irrelevant shortcuts and data leakage. 
Second, item properties such as difficulty and discrimination, as well as the latent constructs they collectively reflect, can be investigated for how they manifest across AI systems, revealing where models systematically succeed or fail.
Together, these two perspectives ground evaluative claims in empirical item-level evidence, supporting more defensible conclusions about capability and deployment.

\textbf{Item-level data enables efficient maintenance, use, and cumulative development of AI benchmarks.}
With increased access to item-level data, score statistics and detailed test cases make it possible to diagnose the validity threats discussed in Sec.~\ref{sec:2.2}, including benchmark saturation and data contamination, in a timely manner.
Item-level analyses can further identify which items have degraded and which remain informative, providing fine-grained guidance for benchmark re-composition, updates, and targeted item generation, saving effort in maintenance rather than multiplying it.
Moreover, a consistent, community-wide release of item-level data lowers the costs of data engineering and allows findings to be reproduced, audited, and compared across studies, enabling cumulative progress and prolonging benchmark lifecycles.
\section{\textsc{OpenEval}: An Item-Centered Benchmark Repository}\label{sec:4}
To show what data infrastructure in our position may look like, we propose an item-centered repository, \textsc{OpenEval}, designed as a community-driven data infrastructure to standardize benchmark items with model responses, metric scores, and other associated information across different tasks, formats, and evaluation harnesses. \textsc{OpenEval} is not the endpoint of our proposal. It is a demonstration that item-level standardization is feasible and a seed for community norms.

\begin{figure}[t]
    \centering
    \includegraphics[width=\linewidth]{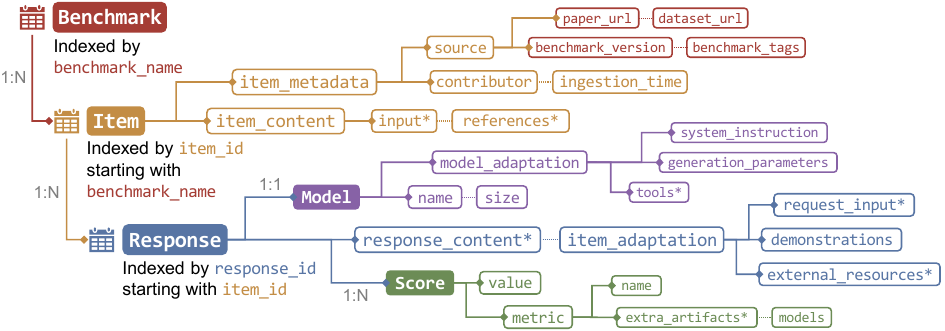}
    \caption{Unified data schema of \textsc{OpenEval}. 
    The left column shows the hierarchical storage structure with indexing keys; 
    the right side expands each entity into its constituent fields. 
    Diamond-tipped connectors point from parent to subfield;  
    dashed connectors indicate folded siblings. 
    1:1 and 1:N denote cardinality, e.g., each response contains exactly one model but may contain multiple scores.
    Asterisks (*) mark fields accepting values in arbitrary formats.}
    \label{fig:schema_main}
\end{figure}

There have been many large-scale, high-quality AI benchmark repositories~(e.g., HELM~\citep{liang2023holistic}, Chatbot Arena~\citep{10.5555/3692070.3692401}, and Open LLM~\citep{open-llm-leaderboard-v2}), some of which release unstructured or semi-structured item-level details alongside the benchmark-level results.
While these resources provide a basis for recent AI evaluation research, the shared data infrastructure remains underdeveloped, hindering cumulative progress across studies and groups.
\textsc{OpenEval} therefore addresses the infrastructure challenge through a \textbf{unified, item-centered schema}, as illustrated on the right side of Fig.~\ref{fig:schema_main} and detailed in App.~\ref{sec:app_schema}, in which each data entry represents a unique item. 

Specifically, the schema is designed around four principles:
\begin{itemize}[itemsep=0pt, topsep=0pt, parsep=1pt,leftmargin=1.5em]
    \item \textit{\textbf{Unified while remaining flexible.}} While effectively standardizing item-level data across different sources, the schema poses relatively loose constraints on fields such as \texttt{response\_content} and \texttt{item\_content} to enable incorporation of heterogeneous content. The typed auxiliary arrays (\texttt{model\_adaptation.tools}, \texttt{item\_adaptation.external\_resources}, and \texttt{metric.extra\_artifacts}) further accommodate items from various evaluation paradigms, including agent and tool-augmented settings, allowing researchers to focus on research opportunities rather than data engineering overhead.
    \item \textit{\textbf{Faithful to evaluation context.}} The schema separates the original item content from the adapted input actually given to the model and records the corresponding model configuration, providing a clean snapshot of the test condition for each response rather than collapsing various experimental conditions into a single record. This self-contained, faithful preservation supports reproducibility and enables meaningful comparisons across different experimental setups.
    \item \textit{\textbf{Informatively rich and versatile.}} The schema archives diverse information at multiple levels, including benchmark-level metadata, test environments, and evaluation metrics, enabling a wide range of practices from provenance tracking and benchmark auditing to coarse- and fine-grained querying by domain, model, or other properties. This richness reflects the paper's core premise: item-level benchmark data serves the science of AI evaluation not for any single analytical goal, but for a wide range of research opportunities.
    \item \textbf{\textit{Scalable for long-term maintenance.}} The unified, flexible, and self-contained nature of each entry naturally supports scalability. As shown on the left side of Fig.~\ref{fig:schema_main}, the repository organizes data into three levels, each indexed by hierarchical identifiers (e.g., \texttt{response\_id} prefixed by \texttt{item\_id}). New benchmarks, items, or model responses can be easily incorporated by appending new entries at the appropriate level without restructuring existing data; this design further facilitates community-driven contributions.
\end{itemize}

So far, to jump-start the community effort, we have been 
(1) collecting evaluation results to supplement item-level data coverage across existing benchmark repositories, 
(2) incorporating interdisciplinary datasets on social aspects of AI, such as \textsc{CulturalBench}~\citep{chiu-etal-2025-culturalbench} and \textsc{EmoBench}~\citep{sabour-etal-2024-emobench},
(3) transforming existing item-level resources from benchmark repositories (e.g., HELM) into \textsc{OpenEval} schema, and
(4) developing a general AI-assisted converter, complemented by an already published data validator, to accommodate formats beyond those supported by harness-specific converters.
\textsc{OpenEval} now covers \textbf{160K+} items across diverse benchmark datasets, with the number of evaluated models per dataset ranging from 11 to 252, resulting in \textbf{10.7M+} item-level responses from 444 distinct models in total, each associated with one or multiple metric scores.

By shifting the norm from reporting aggregate scores to releasing item-level results as a standard part of research publication, we hope that \textsc{OpenEval} will grow into a shared data foundation for the science of AI evaluation and responsible AI deployment, leading us toward more rigorous, empirically grounded validation of benchmark-related claims.
Beyond the technical infrastructure, we envision community governance mechanisms for communicating item usage history, release status, and contamination reports, alongside shared norms for responsible reuse and auditing (App.~\ref{app:governance}).

\section{Empirical Illustrations of Item-Level Analysis}\label{sec:5}
To demonstrate the unique insights enabled by item-level benchmark data, we present illustrative analyses on selected item-level data from \textsc{OpenEval} that provide finer-grained understanding of existing AI benchmark datasets, specifically item quality and construct alignment.

\subsection{Background and Analytical Approaches}
Item-level benchmark analysis has drawn increasing attention in AI evaluation research, with studies examining properties such as item difficulty (e.g., \citep{yao2025the,li-etal-2025-item}), item discrimination, namely the ability to separate different capability levels (e.g., \citep{heineman2025signal,li2025from}), item diversity (e.g., \citep{muennighoff-etal-2023-mteb, yao2025the}), inter-benchmark agreement (e.g., \citep{perlitz2025benchmark, liu2025evaluation}), and downstream performance predictability (e.g., \citep{bhojanam2025prompt, magnusson2025datadecide, wu2025mapping}). As discussed in Sec.~\ref{sec:2.2}, these analyses are typically isolated and difficult to reproduce due to the lack of a large-scale, consistent item-level data release, leaving substantial room for further exploration.
Notably, these studies, to varying degrees, draw on psychometric methodology, which offers a well-established toolkit for item-level analysis grounded in decades of test development and validation practice.
Two types of analysis from this discipline are particularly relevant to AI evaluation. 

The first is item characteristic analysis, which examines whether individual items function as intended with adequate precision for their intended use~\citep{AERA_APA_NCME_2014}; \textit{Classical Test Theory (CTT)} and \textit{Item Response Theory (IRT)} are two foundational frameworks for this purpose~\citep{CookPitoniak2025,hambleton1993comparison}. 
CTT measures item difficulty and discrimination directly from observed score matrices with minimal assumptions, making it readily applicable to large-scale, heterogeneous AI benchmark data~\citep{gulliksen1950theory,lord1968statistical};
IRT provides a more sophisticated, model-based alternative with stronger statistical properties, but has parametric assumptions about the relationship between model ability and item-level performance, with unidimensional formulations requiring careful validation that item responses are driven primarily by a single latent ability~\citep{lord1980applications, hambleton1991fundamentals}.
The second is \textit{Item Factor Analysis (IFA)}, which provides evidence about a test’s internal structure~\citep{anderson1958introduction,reckase200618}. 
The Standards for Educational and Psychological Testing~\citep{AERA_APA_NCME_2014} emphasizes that such structural evidence should be provided when a single aggregate score is interpreted as a meaningful summary of examinee ability.
In the context of AI evaluation, IFA examines whether a benchmark behaves as a coherent measure of the intended model capability versus reflecting construct-irrelevant variance, such as formatting artifacts, label distribution bias, and data contamination.

\begin{figure}[t]
    \centering
    \includegraphics[width=\linewidth]{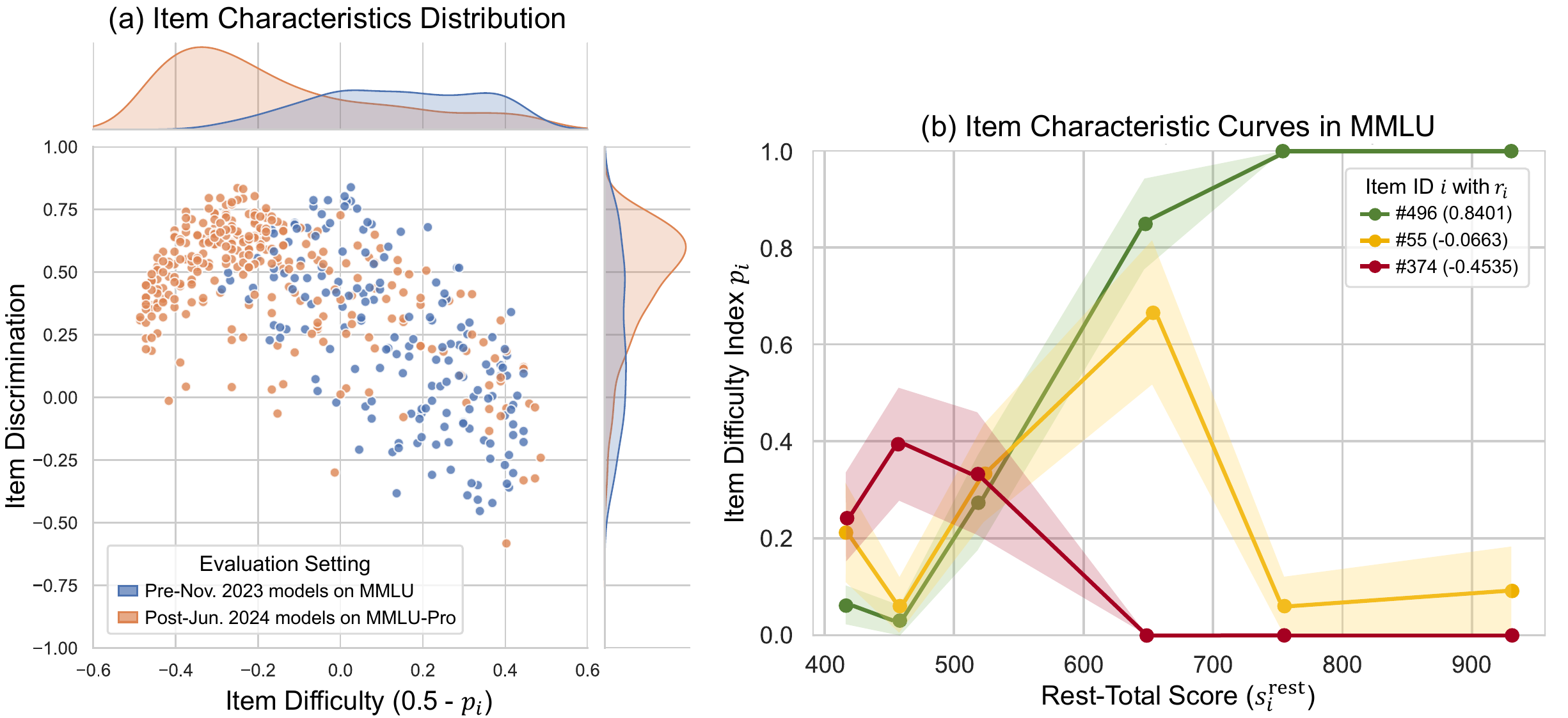}
    \caption{(a) Item characteristics distribution for items from MMLU and \textsc{MMLU-Pro}. Higher item difficulty values correspond to harder items. (b) ICCs for three items in \textsc{MMLU}.}
    \label{fig:ctt}
\end{figure}

\subsection{Examining Item Quality via Classical Test Theory}\label{sec:5.2}

An item's statistical characteristics such as difficulty and discrimination are routinely examined in psychometric test development for quality assurance. 
We conduct a CTT analysis of item characteristics from (1) 66 pre-Nov. 2023 models on 567 items in \textsc{MMLU} \citep{hendrycks2021measuring} and (2) 72 post–June 2024 models on 1,000 items in \textsc{MMLU-Pro} \citep{wang2024mmlupro}. \textsc{MMLU-Pro} \citep{wang2024mmlupro} is an enhanced variant of \textsc{MMLU} intended to increase difficulty and reduce noise via additional distractors, more careful item curation, and expert item review. CTT item analysis provides an empirical way to evaluate these design claims.

For the $i$-th item in a benchmark, 
the \textit{Item Difficulty Index} ($p_i$) can be estimated as~\citep{gulliksen1950theory} 
the proportion of the maximum score achieved on the item averaged across models; a larger $p_i$ indicates an easier item.
The \textit{Item Discrimination} ($r_i$) is measured by the Pearson correlation between the item score and the \textit{Rest-Total Score} ($s^\text{rest}_i$, sum score on all items except $i$) across all measured models~\citep{Henrysson_1963}.
Higher $r_i$ indicates that item $i$ can effectively differentiate models with strong vs. weak overall performance on the benchmark, whereas negative or near-zero $r_i$ suggests a potentially problematic item.

Fig.~\ref{fig:ctt}(a) shows the distributions of item difficulty and discrimination under the two evaluation settings. (Note that the CTT item difficulties on \textsc{MMLU} and \textsc{MMLU-Pro} are specific to their respective sample of models and are not comparable.) 
There are two notable observations: 
(1) The high density of orange observations on the left indicates that \textit{a substantial proportion of \textsc{MMLU-Pro} items have very low difficulty for current models}. In other words, many items are no longer challenging for the 72 post-June 2024 models, suggesting fast benchmark saturation and the need to accelerate benchmark updates. 
(2) Compared to \textsc{MMLU}, \textit{\textsc{MMLU-Pro} exhibits a smaller proportion of items with low discrimination in the evaluated samples}. 
This empirical observation is aligned with \textsc{MMLU-Pro} designers' goal \citep{wang2024mmlupro} to build a more robust, less noisy benchmark.
However, some \textsc{MMLU-Pro} items exhibit poor empirical discrimination despite expert review, warranting further scrutiny for ambiguity, miskeying, construct-irrelevant cues, or measurement of sub-constructs underrepresented in the total score. 

To assess the robustness of these comparisons, we conducted supplementary sensitivity analyses in App.~\ref{sec:app_52}.
First, bootstrap resampling and leave-one-developer-out analysis using an expanded pool of 122 models evaluated on the same 1,000 \textsc{MMLU-Pro} items indicate that the item difficulty and discrimination estimates remain relatively stable across the examined changes in model population;
second, a difficulty-stratified comparison shows that the observed discrimination advantage of \textsc{MMLU-Pro} persists in most difficulty bins, although differences between the evaluated model samples remain a potential confound.

To further understand how item discrimination manifests, we plot the \textit{Item Characteristic Curves} (ICCs) of three items in \textsc{MMLU}. 
For each item $i$, all respondent models are sorted into six equally-sized bins based on their rest-total scores ($s^\text{rest}_i$), representing six levels of overall performance. 
The item difficulty index ($p_i$) within each bin is then plotted in Fig.~\ref{fig:ctt}(b). 
Intuitively, a high-discrimination item's $p_i$ should increase with $s^\text{rest}_i$, meaning that the item appears easier for higher-performing models, resulting in an overall increasing ICC. 
This is the case for item \#496 which had a high item discrimination ($r_{496}$=0.84), whereas for items \#55 and \#374 with near-zero or negative $r_i$, some models that did well on the rest of the benchmark did worse on these items. 

These findings can help identify low-quality items in the benchmark, thereby improving the alignment between benchmark outcomes and design goals at a finer granularity. 
In this sense, detailed examination of item characteristics supports more targeted benchmark maintenance and item generation without requiring wholesale replacement.

\subsection{Revealing Construct Alignment via Item Factor Analyses}\label{sec:5.3}

\begin{figure}[ht]
\begin{minipage}{0.55\textwidth}
    \centering
    \small
    \captionof{table}{Example \textsc{bAbI} item and item counts by reference answers for each cluster in Fig.~\ref{fig:babiqa}.}
    \label{tab:babiqa}
    \begin{tabular}{cccc}
    \toprule
    \multicolumn{4}{c}{\textbf{Example: Item \#1295}} \\
    \midrule
    \multicolumn{4}{l}{Sheep are afraid of mice. Cats are afraid of mice. Jess-} \\
    \multicolumn{4}{l}{ica is a sheep. Wolves are afraid of mice. Mice are afr-} \\
    \multicolumn{4}{l}{aid of wolves. Emily is a wolf. Gertrude is a wolf.} \\
    \multicolumn{4}{l}{Winona is a mouse. Question: What is Emily afraid of?} \\
    \multicolumn{4}{c}{\textbf{Answer}: mouse} \\
    \toprule
    & \multicolumn{3}{c}{\textbf{Number of Items}} \\
    \cmidrule{2-4}
    \textbf{Item Answer} & Cluster \#0 & Cluster \#1 & Cluster \#2 \\
    \midrule
    Sheep & 0 & 0 & 240 \\
    Mouse & 225 & 0 & 0 \\
    Cat & 221 & 3 & 0 \\
    Wolf & 6 & 326 & 0 \\
    \bottomrule
    \end{tabular}
\end{minipage}
\hfill
\begin{minipage}{0.4\textwidth}
    \centering    \includegraphics[width=\linewidth]{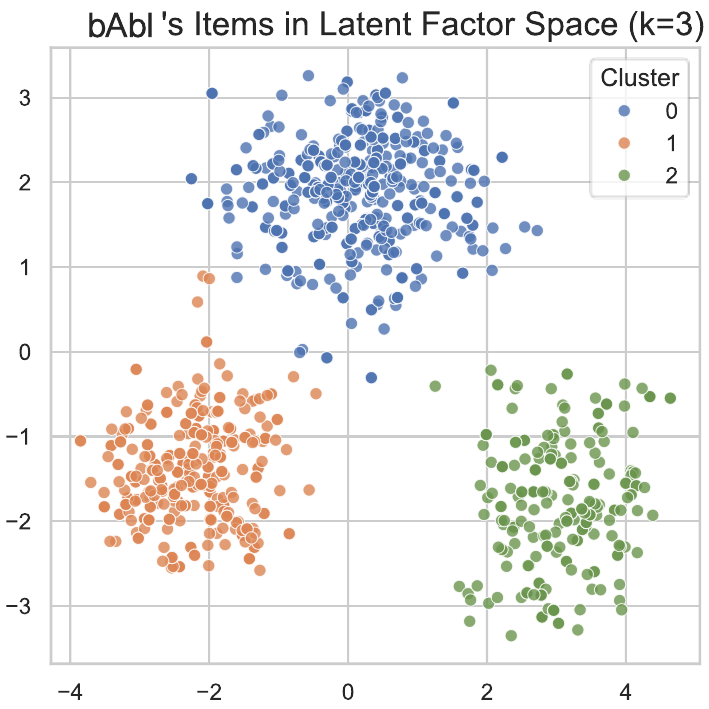}
    \caption{Item clusters on \textsc{bAbI} Task 15 based on factor loadings.}
    \label{fig:babiqa}
\end{minipage}
\end{figure}

As discussed in Sec.~\ref{sec:2.1}, examination of a benchmark's internal structure is essential for understanding whether it measures the intended capabilities or irrelevant confounders. 
Here, we perform variants of conventional IFA for high-dimensional data based on \textit{Singular Value Decomposition} \citep[\textit{SVD};][]{zhang2020note} and \textit{Generalized Low Rank Models} \citep[\textit{GLRM};][]{udell2016generalized}. 
We report findings from \textsc{bAbI} Task 15 (basic deduction) and \textsc{MMLU-Pro}. Additional analysis results on \textsc{MMLU} are presented in App.~\ref{sec:app_53}.

\textsc{bAbI} Task 15~\citep{weston2015towards} aims to assess basic deductive reasoning via inheritance of properties. An example item is shown in Table \ref{tab:babiqa}. 
To aid interpretation, we perform K-means clustering on items' factor loadings on the top 3 factors obtained from SVD-based IFA, with items within each cluster potentially measuring similar sub-constructs. 
As shown in Fig.~\ref{fig:babiqa}, 1,000 items in \textsc{bAbI} form three distinct clusters. A closer examination raises a construct validity flag: Table~\ref{tab:babiqa} shows that item clusters are explained by the reference answer to the item.
This finding demonstrates that different models' performance on \textsc{bAbI} is partially explained by models' propensity to \textit{select specific animals that one is afraid of} (e.g., potentially based on common sense if a model tends to select ``wolf''), rather than the intended \textit{basic deduction capability}, \textit{suggesting potential construct misalignment in \textsc{bAbI} Task 15}.
GLRM-based IFA yields consistent findings.

\begin{wrapfigure}{r}{0.5\textwidth}
    \centering    \includegraphics[width=\linewidth]{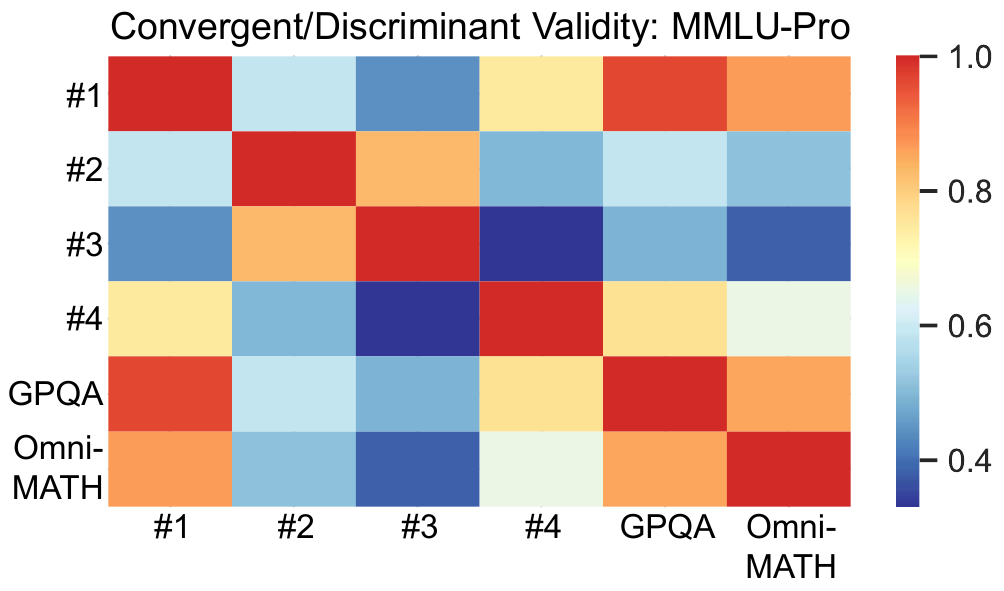}
    \caption{Convergent/discriminant evidence of the four sub-constructs (\#1 - \#4) on \textsc{MMLU-Pro}.}
    \label{fig:convergent_mmlupro}
\end{wrapfigure}

For \textsc{MMLU-Pro}, we interpret the top four factors retained from GLRM-based IFA after varimax rotation. The top 100 items with the largest absolute loadings on each factor are sent to GPT-5 for interpretation. Table \ref{tab:factors_mmlupro} presents a representative item and a sub-construct description for each factor.
The four primary dimensions that best explain differences in model performance appear to reflect different higher-level reasoning capabilities, rather than subject domain proficiency. 
This empirical finding \textit{supports \textsc{MMLU-Pro}’s stated motivation to increase reasoning demands} \citep{wang2024mmlupro} relative to \textsc{MMLU}. 
Indeed, items within the same subject domain (e.g., Psychology and Physics in Fig.~\ref{fig:items}) could differ substantially in loadings on the four factors.

As an external plausibility check using convergent and discriminant validity evidence \citep{campbell1959convergent}, we correlate factor subscores (mean scores on the 100 items with the largest absolute loadings) with scores on two external benchmarks: \textsc{GPQA}~\citep{rein2024gpqa} (graduate-level biology, physics, and chemistry) and \textsc{Omni-MATH}~\citep{gao2025omnimath} (Olympiad-level mathematics). Both target high-level formal reasoning, with the former grounded more in applied scientific contexts. We hypothesize that \textit{Factor \#1 (formal, quantitative, multi-step modeling) aligns with both formal reasoning benchmarks}, whereas \textit{Factor \#4 (applied synthesis and case-based judgment) aligns more with \textsc{GPQA}}. Results in Fig.~\ref{fig:convergent_mmlupro} are broadly consistent with these hypotheses. Further, Factors \#2 (domain-specific recall and simple reasoning) and \#3 (conceptual understanding and explanation) show weaker correlations with both \textsc{GPQA} and \textsc{Omni-MATH}, providing discriminant validity evidence. We treat these findings as descriptive rather than definitive evidence supporting these sub-construct interpretations.

\begin{table*}
    \centering
    \footnotesize
    \caption{Representative items with large absolute item factor loadings and possible constructs for each GLRM factor in \textsc{MMLU-Pro}. The top 100 representative items are interpreted and summarized into a candidate label by GPT-5, then manually revised.}
    \label{tab:factors_mmlupro}
    \begin{tabular}{cll}
    \toprule
    \textbf{Factor} & \textbf{Representative Item} & \textbf{Potential Sub-Construct} \\
    \midrule
    \multirow{4}{*}{MMLU-Pro \#1} & A 10kVA, 2400/240V, single-phase transformer has the following &\\
    & resistances and leakage reactance. Find the primary voltage & Formal, quantitative, \\
    & required to produce 240V at the secondary terminals at full load, & multi-step modeling\\
    & when the load power factor is 0.8 power factor lagging/leading. & \\
    \midrule
    \multirow{2}{*}{MMLU-Pro \#2} & What are the principal and selective instruments of control of which & Domain-specific recall\\
    & the Federal Reserve System makes use? & and simple reasoning\\
    \midrule
    \multirow{2}{*}{MMLU-Pro \#3} & \multirow{2}{*}{What is meant by the term ``hypothesis testing''?} & Conceptual understanding\\
    && and explanation \\
    \midrule
    & Which of the following might explain how a price decrease might & \multirow{2}{*}{Applied synthesis and}\\
    MMLU-Pro \#4 & cause a decrease in quantity demanded and an upward-sloping & \multirow{2}{*}{case-based judgment}\\
    & demand curve? & \\
    \bottomrule
    \end{tabular}
\end{table*}



\section{Broader Implications beyond Benchmarking}\label{sec:6}
The benefits of releasing item-level data extend beyond AI evaluation.
Shared infrastructure of this kind could benefit AI research, ground policy in evidence, and democratize AI through participatory evaluation.

\textbf{For AI Researchers.}
Releasing such data at scale could contribute to ML theory, where central questions about generalization and learning dynamics are inherently item-level. Methods built on item-level data have advanced principled training: identifying which examples drive learning~\citep{swayamdipta-etal-2020-dataset}, informing data mixing~\citep{10.5555/3692070.3692443,xie2023doremi}, and grounding curriculum learning~\citep{10.1145/1553374.1553380} through measurable item difficulty. 
The same per-instance signal supports estimating generalization gaps and attributing predictions to training examples~\citep{xu-etal-2024-position}. 

\textbf{For Policy Makers.}
Policy and governance decisions increasingly rely on AI benchmark results to justify claims about model capability, risk, and deployment readiness (e.g.,~\cite{anderljung2023frontier,10.1145/3708359.3712152,eu2024aiact}). 
Item-level data lets policy makers inspect what a benchmark actually measures, enabling contextualized judgments about how much weight a score should carry for a given policy question. The evaluation methods built on such data further equip regulators to audit benchmark design, verify developer claims, and ground policy decisions in independently checkable evidence.

\textbf{For End Users.}
The communities affected by AI deployment are largely absent from the evaluations that justify it. Item-level responses, not just items, make user engagement substantive: users can see whether a model actually handles cases like theirs, detect performance gaps that aggregate scores hide, and compare options before deployment \cite{10.1145/3351095.3372873}. 
Interfaces built on this infrastructure can lower the technical barrier to participation, making participatory evaluation practical: evaluation as a channel through which users help shape priorities, and through them, model development \cite{liao2025rethinking}.

\section{Alternative Views}\label{sec:7}
\textit{Item-level release worsens contamination.} 
A natural objection is that releasing item-level data increases data contamination risk~\cite{sculley2025position, jacovi-etal-2023-stop}. 
We argue that openness and contamination prevention need not be contradictory.
Information asymmetry about what benchmark items exist and how they have been used complicates both prevention and detection.
Standardized releases can help developers identify and filter known benchmark content from training data, while item-level responses and test conditions provide evidence for independent scrutiny of decontamination claims and investigation of suspected contamination.
Lowering the cost of external auditing can also help hold developers accountable for undisclosed benchmark use.
In addition, our position is compatible with hold-out designs, verified access, and staged release after a fixed embargo period, allowing disclosure practices to account for contamination risks and licensing restrictions.
We believe the benefits of independently verifiable evaluation outweigh the potential contamination risks, provided that releases incorporate appropriate safeguards.

\textit{Mandating item-level release creates additional burden.}
Standardized item-level release does require additional effort from benchmark authors and model evaluators. 
However, because model responses are already generated during evaluation, releasing retained results does not require rerunning inference.
The additional work primarily involves conversion, validation, quality control, and uploading.
As illustrated by \textsc{OpenEval}, shared infrastructure can reduce duplicated effort: contributors submit to a common archive, a unified schema specifies what to release and in what format, and converters for widely used evaluation harnesses reduce the reformatting burden.
The cost of individual releases must nevertheless be distinguished from the collective cost of pooling data for analyses across models and benchmarks.
Soliciting contributions, coordinating participants, expanding response coverage, and maintaining storage and access require sustained community effort.
Standardization makes individual releases easier to combine, reducing the need for each research group to independently assemble and preprocess the same data.
Contributors can, in turn, benefit from the accumulated contributions of others, creating incentives for further participation.

\begin{ack}
This work was supported by Schmidt Sciences. 
\end{ack}

\bibliographystyle{abbrv}
\bibliography{example_paper}

\appendix
\section{Limitations}\label{sec:limit}
The limitations discussed here pertain to how we present and substantiate our position in this paper, rather than to the value or potential of item-level benchmark data itself.

First, the empirical illustrations in Sec.~\ref{sec:5} draw primarily on classical test theory and item factor analysis, representing only a subset of the analytical possibilities enabled by item-level data.
Applying psychometric methods to AI benchmarking data also requires caution, as their underlying assumptions may not hold: shared training corpora and model lineages can introduce dependencies among models, while benchmark multidimensionality can complicate the interpretation of rest-total discrimination.
Although our sensitivity analyses (App.~\ref{sec:app_52}) support the stability of CTT estimates, they do not establish model independence.
The GPT-assisted factor interpretations in Sec.~\ref{sec:5.3} also remain exploratory.
Broader methodological coverage, including item response theory, differential item functioning analysis, and diagnostic classification models, would further demonstrate the range of research practices that item-level data can support.

Second, although Secs.~\ref{sec:2}, \ref{sec:4}, \& \ref{sec:7} discuss the tension between open data access and data contamination risk, and App.~\ref{app:governance} elaborates potential safeguards and governance mechanisms, their effectiveness and long-term sustainability remain to be established in practice.
Implementing and evaluating access controls, staged releases, and community reporting processes will be necessary to assess how well these mechanisms balance transparency, reproducibility, privacy, and benchmark integrity.

\section{Data Accessibility and Governance}\label{app:governance}

Standardized item-level releases must balance transparency and reproducibility with privacy, licensing restrictions, and benchmark integrity.
Our position advocates maximizing access whenever legally and ethically feasible, while allowing release mechanisms to accommodate benchmark-specific constraints.
Here, we elaborate how standardized releases can support contamination prevention and auditing, and outline complementary safeguards and community governance mechanisms.

\paragraph{Supporting Contamination Prevention and Auditing.}
Information asymmetry about which benchmark items exist, how they have been used, and what uses are permitted complicates contamination prevention and detection.
Standardized item content and provenance records can help developers identify known benchmark material and filter it from training data.
Item-level responses and test conditions further support independent scrutiny of decontamination claims by enabling researchers to investigate item-level contamination signals.
Such signals warrant further investigation rather than conclusive evidence of contamination or its absence.
Making such evidence accessible also lowers the cost of developing external auditing methods and investigating undisclosed benchmark use.

\paragraph{Controlled Release Mechanisms.}
\begin{itemize}[itemsep=0pt, topsep=0pt, parsep=1pt,leftmargin=1.5em]
    \item \textit{Hold-out designs:}
    A portion of benchmark items can be withheld from public release and reserved for independent, periodic re-evaluation.
    Comparisons between performance on public and held-out items can help investigate potential overfitting or contamination, provided that differences in item difficulty and content are taken into account.
    This design preserves a protected evaluation resource while allowing the released portion to support item-level analysis and methodological development.

    \item \textit{Verified access:}
    For content that cannot be released publicly but can be shared under appropriate conditions, access can be restricted to researchers who register and agree to specified usage terms, with additional identity or affiliation verification where warranted.
    Such arrangements can specify permitted uses, redistribution restrictions, and reporting obligations.
    Access remains subject to the underlying privacy and licensing constraints; registration alone does not remove these restrictions.

    \item \textit{Staged release:}
    Items can be released after a predefined embargo period, such as one evaluation cycle, to balance timely community access with the value of temporarily protected test content.
    The release schedule and access conditions should be documented in advance so that contributors and users can track when and how the data will become available.
    Content that remains subject to privacy or licensing restrictions would continue to require appropriate access controls after the embargo ends.
\end{itemize}

\paragraph{Community Governance and Long-Term Sustainability.}
We envision a coordinated community process for maintaining item-level data standards and promoting transparency. 
A shared communication channel would allow contributors and users to report known benchmark usage, staged-release status, and suspected contamination, supported by community norms for disclosure and auditing. 
Reports should distinguish suspected from substantiated contamination and allow supporting evidence and corrections to be recorded, reducing information asymmetry without claiming complete visibility into undisclosed data use. 
Regarding the schema governance, a lead organizing team would coordinate schema updates based on community feedback, with version control and backward-compatibility requirements to limit fragmentation. 
Harness-specific converters, general conversion tools, and schema validation can reduce submission errors, while additional quality control is needed to ensure content accuracy, provenance, and metadata consistency. 
Long-term sustainability also requires resources for storage, access, version management, corrections, and contributor support. 
Drawing on practices from established repositories, we are exploring federated governance under common standards and central coordination. 
Shared infrastructure can distribute responsibilities and reduce duplicated effort, but sustained funding and maintenance remain necessary.

\section{OpenEval Schema}\label{sec:app_schema}
\begin{figure*}[!htbp]
    \centering
    \includegraphics[width=\linewidth]{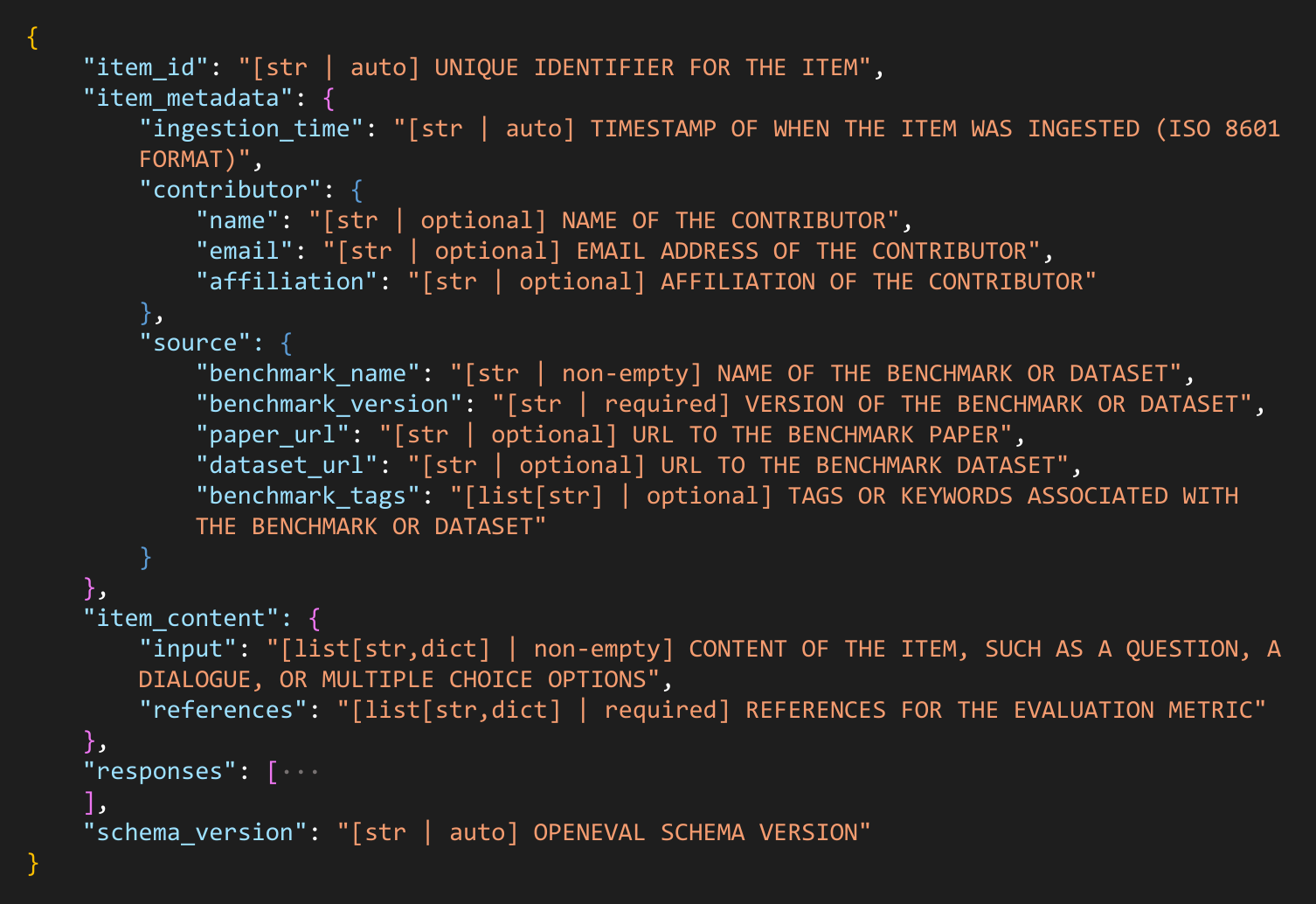}
    \caption{\textsc{OpenEval} data schema (item). The value of each field specifies the data type, presence requirement, and a short explanation of the field. Specifically, \texttt{auto} means the field is auto-generated, \texttt{optional} means the field can be absent from the data entry, \texttt{required} means the data contributors are required to specify this field, and \texttt{non-empty} means the field cannot be left blank.}
    \label{fig:schema}
\end{figure*}

\begin{figure*}[!htbp]
    \centering
    \includegraphics[width=\linewidth]{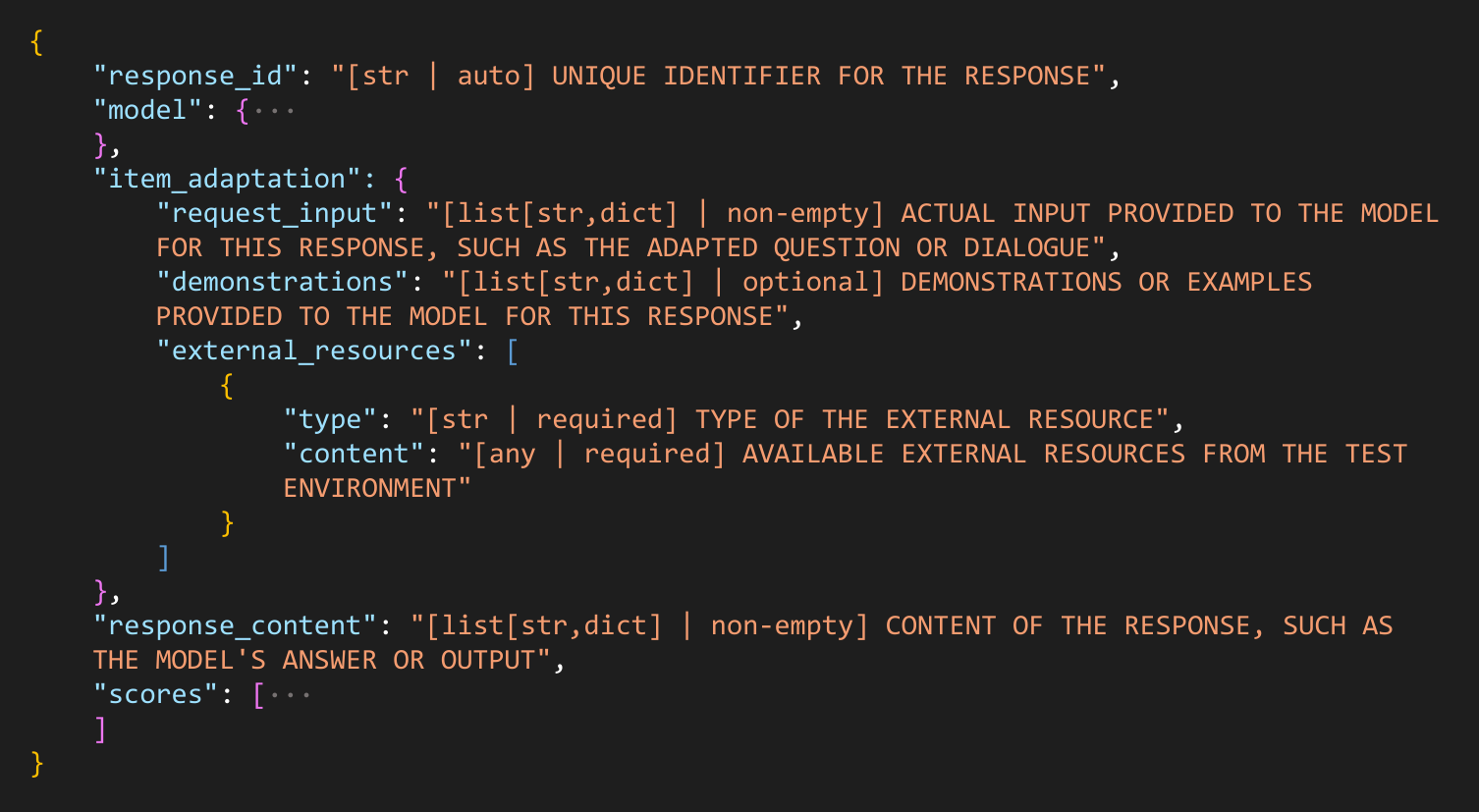}
    \caption{\textsc{OpenEval} data schema (response). Each response object is an element of the \texttt{responses} field in Fig.~\ref{fig:schema}.}
    \label{fig:schema_res}
\end{figure*}

\begin{figure*}[!htbp]
    \centering
    \includegraphics[width=\linewidth]{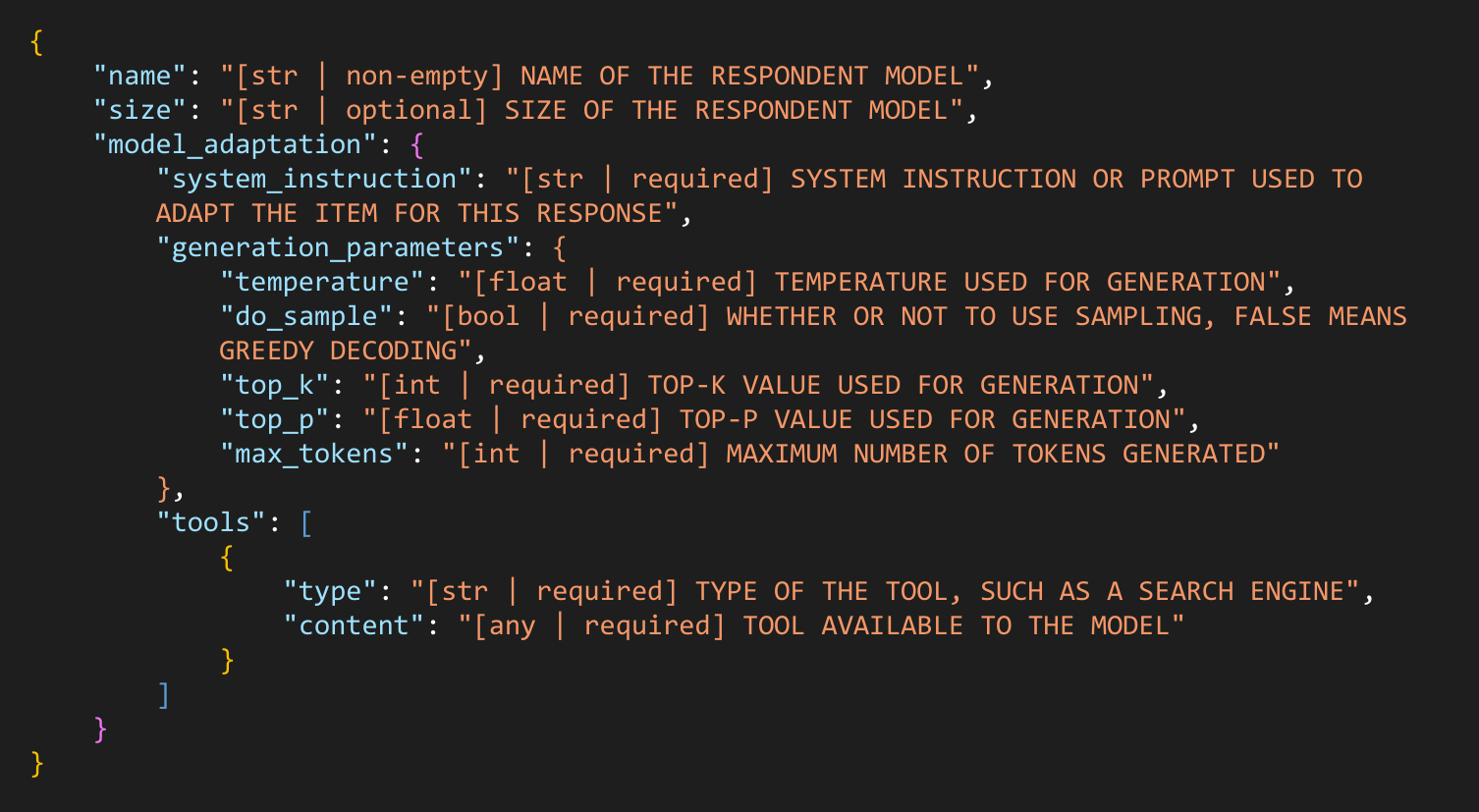}
    \caption{\textsc{OpenEval} data schema (model). The model object here corresponds to the \texttt{model} field in Fig.~\ref{fig:schema_res}.}
    \label{fig:schema_model}
\end{figure*}

\begin{figure}[!htbp]
    \centering
    \includegraphics[width=\linewidth]{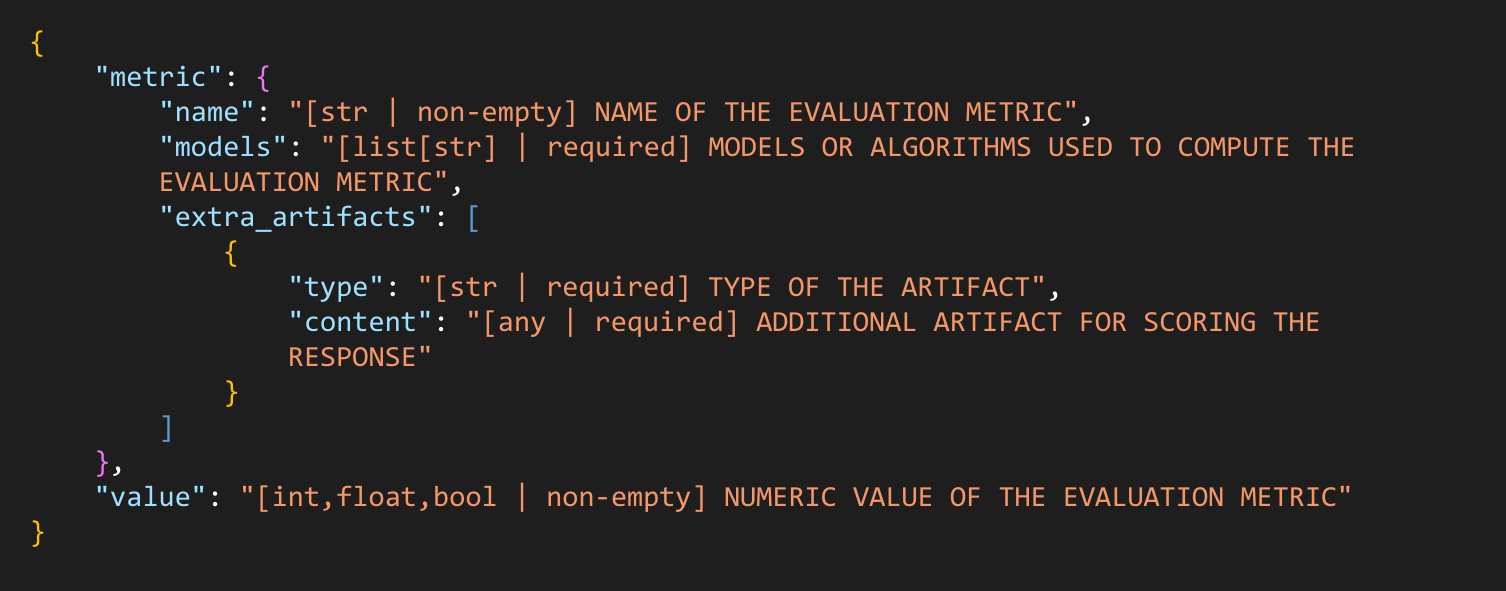}
    \caption{\textsc{OpenEval} data schema (score). Each score object is an element of the \texttt{scores} field in Fig.~\ref{fig:schema_res}.}
    \label{fig:schema_score}
\end{figure}

Fig.~\ref{fig:schema} displays the data schema of OpenEval, including required and optional fields, data type constraints, and field descriptions. The \texttt{response}, \texttt{model}, and \texttt{score} objects folded in Fig.~\ref{fig:schema} are detailed in Figs.~\ref{fig:schema_res}, \ref{fig:schema_model}, and \ref{fig:schema_score}, respectively.

\FloatBarrier
\section{Sensitivity Analyses of CTT Estimates}\label{sec:app_52}

To assess the sensitivity of the item difficulty and discrimination estimates to the evaluated model population, we expanded the original sample of 72 models to 122 models evaluated on the same 1,000 \textsc{MMLU-Pro} items. 
The original sample spans 14 developers and 20 model families, while the expanded sample covers 17 developers. 
We conducted two complementary analyses: bootstrap resampling and leave-one-developer-out analysis.

\paragraph{Bootstrap Resampling.}
We generated 1,000 bootstrap samples, each by drawing 72 models with replacement from the expanded pool of 122 models, and re-estimated item difficulty and discrimination in each sample. Table~\ref{tab:ctt_bootstrap} reports the mean item-level standard error and 90\% confidence interval width, together with the mean correlation across items between the resampled estimates and those obtained from the original 72-model sample. The average correlations were 0.9916 for difficulty and 0.9732 for discrimination, indicating that the relative ordering of item estimates remained largely stable under the examined changes in model composition.

\paragraph{Leave-one-developer-out Analysis.}
To further examine whether the estimates were disproportionately influenced by any single developer, we excluded all models from each of the 17 developers in turn and re-estimated item difficulty and discrimination using the remaining models. 
For each exclusion, we computed the Pearson correlation across items between the resulting estimates and the full 122-model estimates, using items with defined estimates in both samples. 
As shown in Table~\ref{tab:ctt_bootstrap}, these correlations averaged 0.9988 for difficulty and 0.9973 for discrimination. 
Table~\ref{tab:ctt_developer} reports the results for the five largest developers; all correlations exceeded 0.988.

\begin{table}[ht]
    \centering
    \caption{
    Sensitivity of CTT estimates on 1,000 \textsc{MMLU-Pro} items.
    The correlation reference is the original 72-model sample for bootstrap resampling and the full 122-model sample for leave-one-developer-out analysis.
    For leave-one-developer-out analysis, SEs and interval widths are jackknife-style descriptive summaries.
    }
    \begin{tabular}{llccc}
        \toprule
        \textbf{Analysis} & \textbf{Estimate} & \textbf{Correlation} & \textbf{SE} & \textbf{90\% CI Width} \\
        \midrule
        \multirow{2}{*}{Bootstrap} & Difficulty & 0.9916 & 0.0523 & 0.1714 \\
        & Discrimination & 0.9732 & 0.0691 & 0.2295 \\
        \midrule
        Leave-one- &  Difficulty & 0.9988 & 0.0718 & 0.2361 \\
        developer-out & Discrimination & 0.9973 & 0.0646 & 0.2127 \\
        \bottomrule
    \end{tabular}
    \label{tab:ctt_bootstrap}
\end{table}

\begin{table}[ht]
\centering
\caption{
    Stability of CTT estimates after excluding each of the five largest model developers. 
    Correlations are computed across items against the full 122-model estimates; 
    model counts indicate the number of models excluded.
    }
\label{tab:ctt_developer}
\begin{tabular}{lccc}
    \toprule
    && \multicolumn{2}{c}{\textbf{Full-Sample Correlation}} \\
    \cmidrule{3-4}
    \textbf{Developer} & \textbf{Model Count} & \textit{Difficulty} & \textit{Discrimination}\\
    \midrule
    Qwen      & 27 & 0.9951 & 0.9899 \\
    OpenAI    & 21 & 0.9936 & 0.9882 \\
    Google    & 19 & 0.9976 & 0.9924 \\
    Anthropic & 10 & 0.9980 & 0.9952 \\
    Meta      & 8  & 0.9993 & 0.9976 \\
    \bottomrule
\end{tabular}
\end{table}

Regarding the jackknife-style approximation in the leave-one-developer-out analysis, we calculated the standard error for each item $j$ as follows:
\begin{equation}
\widehat{\mathrm{SE}}_j
=
\frac{D-1}{\sqrt{D}}\,
\operatorname{SD}_{d=1,\ldots,D}
\left(\hat{\theta}_{j,-d}\right),
\end{equation}
where $\hat{\theta}_{j,-d}$ denotes the item's difficulty or discrimination estimate after excluding developer $d$, $D=17$, and $\operatorname{SD}$ is the sample standard deviation. 
The approximate 90\% CI width was then computed as $2 \times 1.645 \times \widehat{\mathrm{SE}}_j$, and both quantities were averaged across items. 
These quantities should be interpreted as descriptive sensitivity summaries rather than calibrated confidence intervals.

\paragraph{Difficulty-Stratified Discrimination Comparison.}
To examine whether the observed discrimination differences between \textsc{MMLU} and \textsc{MMLU-Pro} persist within similar ranges of observed item difficulty, we grouped items by their difficulty index $p_i$ into five bins of width 0.2 and compared item discrimination within each bin using Mann--Whitney U tests. 
We excluded 22 zero-variance \textsc{MMLU-Pro} items for which discrimination was undefined.

As shown in Table~\ref{tab:ctt_difficulty_bins}, \textsc{MMLU-Pro} exhibits higher mean discrimination in all five bins. 
Among the first four bins, three show statistically significant differences ($p<0.01$), while the difference in $[0.4,0.6)$ is not significant ($p=0.333$). 
The final bin contains only two \textsc{MMLU} items, limiting the interpretability of that comparison. 
These results suggest that the observed discrimination advantage is not solely attributable to differences in the overall distribution of observed item difficulty. 

\begin{table}[ht]
    \centering
    \caption{Item discrimination within difficulty-index bins for \textsc{MMLU} and \textsc{MMLU-Pro}. A larger difficulty-index $p_i$ indicates an easier item. The $p$-values are from within-bin Mann--Whitney U tests; 22 zero-variance \textsc{MMLU-Pro} items were excluded.}
    \begin{tabular}{clccc}
        \toprule
        \textbf{$p_i$ Bin} & \textbf{Dataset} & \textbf{Bin Size} & \textbf{Mean Disc.} & \textbf{$p$-value} \\
        \midrule
        \multirow{2}{*}{$(0.0,0.2)$} & \textsc{MMLU} & 144 & -0.0174 & \multirow{2}{*}{$2.48\times10^{-3}$} \\
        & \textsc{MMLU-Pro} & 86 & 0.0615 & \\
        \midrule
        \multirow{2}{*}{$[0.2,0.4)$} & \textsc{MMLU} & 172 & 0.1444 & \multirow{2}{*}{$3.43\times10^{-6}$} \\
        & \textsc{MMLU-Pro} & 86 & 0.2933 & \\
        \midrule
        \multirow{2}{*}{$[0.4,0.6)$} & \textsc{MMLU} & 175 & 0.4161 & \multirow{2}{*}{$3.33\times10^{-1}$} \\
        & \textsc{MMLU-Pro} & 152 & 0.4708 & \\
        \midrule
        \multirow{2}{*}{$[0.6,0.8)$} & \textsc{MMLU} & 74 & 0.4605 & \multirow{2}{*}{$7.16\times10^{-13}$} \\
        & \textsc{MMLU-Pro} & 281 & 0.5876 & \\
        \midrule
        \multirow{2}{*}{$[0.8,1.0)$} & \textsc{MMLU} & 2 & 0.3803 & \multirow{2}{*}{$1.18\times10^{-1}$} \\
        & \textsc{MMLU-Pro} & 373 & 0.4967 & \\
        \bottomrule
    \end{tabular}
    \label{tab:ctt_difficulty_bins}
\end{table}

Together, these analyses indicate that the CTT estimates, particularly their relative ordering across items, remain relatively stable and are not disproportionately driven by any single developer. 
These findings do not establish independence among models or isolate benchmark design effects from differences between the model samples evaluated on \textsc{MMLU} and \textsc{MMLU-Pro} and thus remain descriptive.

\section{Additional IFA Results}\label{sec:app_53}
\begin{figure*}[h]
    \centering
    \includegraphics[width=0.6\linewidth]{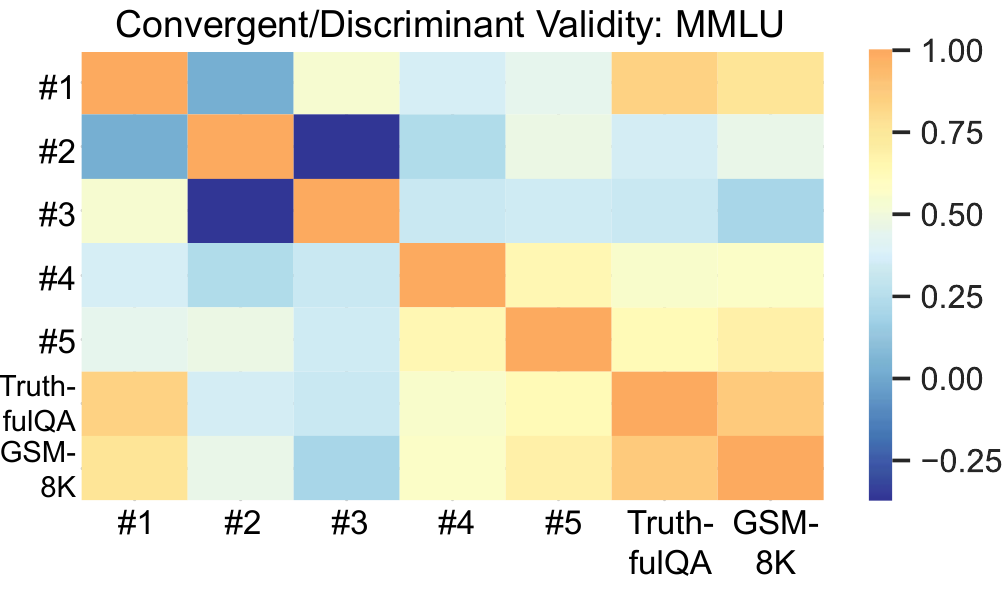}
    \caption{Convergent/discriminant evidence of the five sub-constructs (\#1 - \# 5) on \textsc{MMLU}.}
    \label{fig:convergent_mmlu}
\end{figure*}

\begin{figure*}[h]
    \centering
    \includegraphics[width=\linewidth]{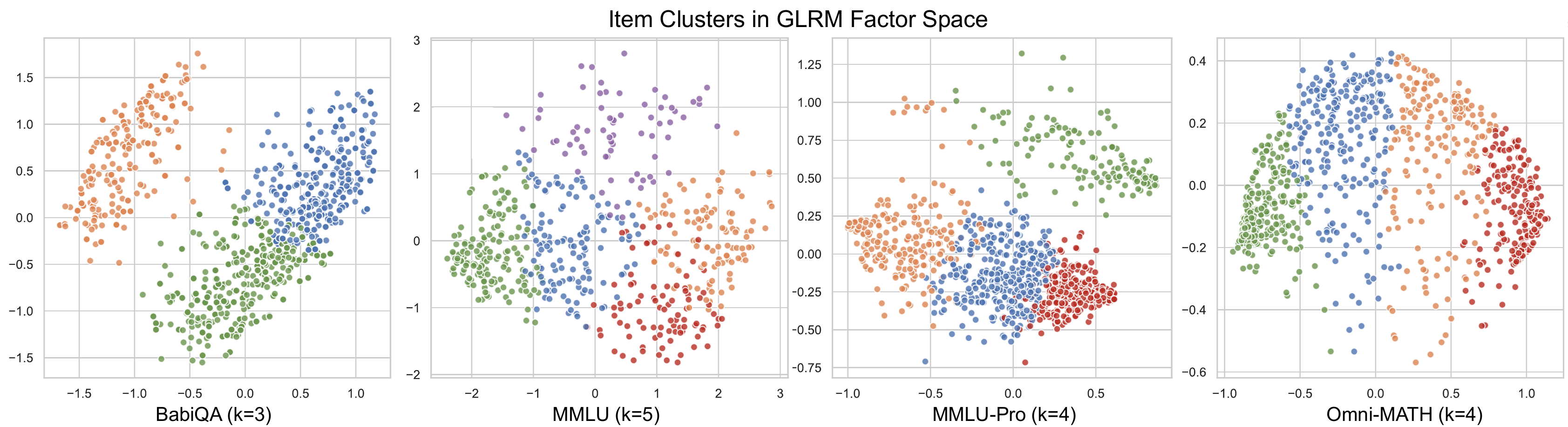}
    \caption{Clusters from four benchmark datasets in HELM revealed by K-means clustering over item factor loadings from GLRM.}
    \label{fig:glrm}
\end{figure*}

\begin{figure*}[t]
    \centering
    \includegraphics[width=\linewidth]{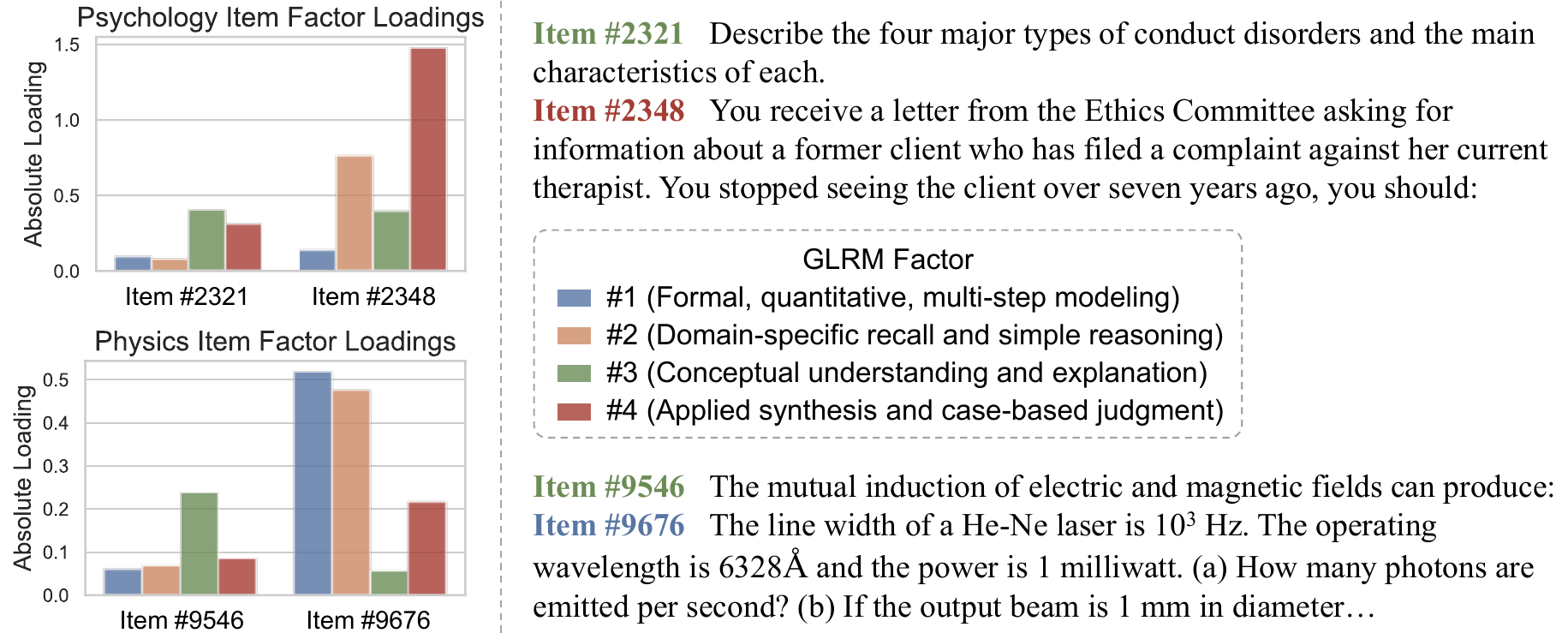}
    \caption{Example items with different maximum factor loadings within the same subject (psychology and physics) in \textsc{MMLU-Pro}.}
    \label{fig:items}
\end{figure*}

Table~\ref{tab:factors_mmlu} and Fig.~\ref{fig:convergent_mmlu} report the results of the GLRM analysis conducted in Sec.~\ref{sec:5.3} on \textsc{MMLU}.
Fig.~\ref{fig:glrm} shows item clusters from four benchmark datasets in HELM, revealed by K-means clustering over item factor loadings derived from GLRM.
Fig.~\ref{fig:items} illustrates that items within the same subject or dataset, despite sharing the same label, can emphasize different aspects when their maximum item factor loadings differ.

\begin{table*}[h]
    \centering
    \small
    \caption{Representative items with large absolute item factor loadings and possible constructs for each GLRM factor in \textsc{MMLU}. The top 100 representative items are interpreted and summarized into a candidate label by GPT-5, then manually revised.}
    \label{tab:factors_mmlu}
    \begin{tabular}{cll}
    \toprule
    \textbf{Factor} & \textbf{Representative Item} & \textbf{Potential Sub-Construct} \\
    \midrule
    \multirow{2}{*}{MMLU \#1} & Which one of the following statements best describes the algebraic  & Domain-specific canonical \\
    & representation of the fitted regression line? & framework knowledge \\
    \midrule
    & Based on the paper “SoK: SSL and HTTPS: Revisiting past & \multirow{2}{*}{Applied synthesis and}\\
    MMLU \#2 & challenges and evaluating certificates trust model enhancements”, & \multirow{2}{*}{case-based judgment }\\
    & which of the following statements are false? &\\
    \midrule
    \multirow{2}{*}{MMLU \#3} & Find the product of the given polynomials in the given polynomial & Formal, quantitative,\\
    & ring. $f(x) = 4x - 5, g(x) = 2x^2 - 4x + 2$ in $\mathbb{Z}_8[x]$. & multi-step modeling\\
    \midrule
    \multirow{2}{*}{MMLU \#4} & Which of the following statements is true concerning the population & Domain-specific recall\\
    & regression function (PRF) and sample regression function (SRF)? & and simple reasoning\\
    \midrule
    & The (\enspace \enspace) is categorized as an unknown segment of the Deep Web & \multirow{2}{*}{Conceptual understanding}\\
    MMLU \#5 & which has been purposely kept hidden and is inaccessible using & \multirow{2}{*}{and explanation}\\
    & standard web browsers. &\\
    \bottomrule
    \end{tabular}
\end{table*}

\end{document}